\documentclass[conference]{IEEEtran}

\usepackage{amsfonts, gensymb, mathtools, microtype}

\usepackage[main=english, ngerman]{babel}
\usepackage[T1]{fontenc}
\usepackage[utf8]{inputenc}

\usepackage[hyphens]{url}
\usepackage[colorlinks, citecolor=LimeGreen]{hyperref}
\usepackage[dvipsnames, table]{xcolor}

\usepackage{tikz}
\usetikzlibrary{
    arrows,
    backgrounds,
    calc,
    decorations,
    calligraphy,
    fit,
    positioning,
    shapes,
    spy
}

\usepackage{pgfplots}
\pgfplotsset{compat=newest}
\usepgfplotslibrary{colorbrewer}

\usepackage[hang]{caption}

\usepackage{array, diagbox, makecell, multirow, subfig, xfrac}

\usepackage[export]{adjustbox}

\usepackage{graphicx, graphbox}
\usepackage[figuresright]{rotating}
\graphicspath{{figs/}}

\usepackage{listings}
\usepackage[defaultlines=3, all]{nowidow}

\newcolumntype{P}[1]{>{ \centering  \arraybackslash }p{#1}}
\newcolumntype{Q}[1]{>{ \raggedleft \arraybackslash }p{#1}}
\newcolumntype{M}[1]{>{ \centering  \arraybackslash }m{#1}}
\newcolumntype{N}[1]{>{ \raggedleft \arraybackslash }m{#1}}
\newcolumntype{B}[1]{>{ \centering  \arraybackslash }b{#1}}
\newcolumntype{C}[1]{>{ \raggedleft \arraybackslash }b{#1}}

\usepackage{ccicons}

\usepackage{packages/hex}

\usepackage{fontawesome5}

\newcommand{\ORCID}[1]{\textsuperscript{\href{https://orcid.org/#1}{\textcolor[HTML]{A6CE39}{\faOrcid}}}}

\newcommand{\ORCIDSchlosser}{0000-0002-0682-4284} % ORCID Tobias Schlosser
\newcommand{\ORCIDFriedrich}{0000-0001-6326-4749} % ORCID Michael Friedrich
\newcommand{\ORCIDKowerko}{0000-0002-4538-7814}   % ORCID Danny Kowerko

\usepackage{fancyhdr, lastpage}
\fancypagestyle{plain}{\fancyfoot[C]{\thepage/\pageref{LastPage}}}
\begin{document}

%%%%%%%%%%%%%%%%%%%%%%%%%%%%%%%%%%%% Title %%%%%%%%%%%%%%%%%%%%%%%%%%%%%%%%%%%%%
\title{Hexagonal Image Processing in the Context of Machine Learning: Conception of a Biologically Inspired Hexagonal Deep Learning Framework}

\author{
    \IEEEauthorblockN{
        Tobias Schlosser\ORCID{\ORCIDSchlosser},
        Michael Friedrich\ORCID{\ORCIDFriedrich}, and
        Danny Kowerko\ORCID{\ORCIDKowerko}
    }
    \IEEEauthorblockA{
        Junior Professorship of Media Computing, \\
        Chemnitz University of Technology, \\
        09107 Chemnitz, Germany, \\
        \texttt{\small \{firstname.lastname\}@cs.tu-chemnitz.de}
    }
}
%%%%%%%%%%%%%%%%%%%%%%%%%%%%%%%%%%%% Title %%%%%%%%%%%%%%%%%%%%%%%%%%%%%%%%%%%%%

\maketitle

\begin{abstract}
    %%%%%%%%%%%%%%%%%%%%%%%%%%%%%%%%%%%%% Text %%%%%%%%%%%%%%%%%%%%%%%%%%%%%%%%%%%%%
    Inspired by the human visual perception system, hexagonal image processing in the context of machine learning deals with the development of image processing systems that combine the advantages of evolutionary motivated structures based on biological models. While conventional state-of-the-art image processing systems of recording and output devices almost exclusively utilize square arranged methods, their hexagonal counterparts offer a number of key advantages that can benefit both researchers and users. This contribution serves as a general application-oriented approach the synthesis of the therefore designed hexagonal image processing framework, called Hexnet, the processing steps of hexagonal image transformation, and dependent methods. The results of our created test environment show that the realized framework surpasses current approaches of hexagonal image processing systems, while hexagonal artificial neural networks can benefit from the implemented hexagonal architecture. As hexagonal lattice format based deep neural networks, also called H-DNN, can be compared to their square counterparts by transforming classical square lattice based data sets into their hexagonal representation, they can also result in a reduction of trainable parameters as well as result in increased training and test rates.
    %%%%%%%%%%%%%%%%%%%%%%%%%%%%%%%%%%%%% Text %%%%%%%%%%%%%%%%%%%%%%%%%%%%%%%%%%%%%
\end{abstract}


\section{Introduction and motivation}

%%%%%%%%%%%%%%%%%%%%%%%%%%%%%%%%%%%%% Text %%%%%%%%%%%%%%%%%%%%%%%%%%%%%%%%%%%%%
Following the developments of recent years, machine learning methods in the form of artificial neural networks are becoming increasingly important. Convolutional neural networks (CNN) for object recognition and classification in this context are one of the approaches that are in the focus of current research in the domain of deep learning. In order to meet the ever-increasing demands of more complex problems, novel application areas, and larger data sets, following \textit{Krizhevsky et al.} \cite{Krizhevsky2012}, more and more novel models and procedures are being developed, as their complexity and diversity following \textit{Szegedy et al.} \cite{Szegedy2015} steadily increases. However, novel architectures for artificial neural networks, including convolutional and pooling layers, are being developed that exceed the conventional cartesian-based architectures. These include, but are not limited to, spherical or non-Euclidean manifolds, as current research by \textit{Bronstein et al.} \cite{Bronstein2017} demonstrates.

While the structure and functionality of artificial neural networks is inspired by biological processes, they are also limited by their underlying structure due to the current state of the art of recording and output devices. Therefore, mostly square structures are used, which also significantly restrict subsequent image processing systems, as the set of allowed operations is reduced depending on the arrangement of the underlying structure \cite{Staunton1990}.

In comparison to the current state of the art in machine learning, the human visual perception system suggests an alternative, evolutionarily-based structure, which manifests itself in the human eye. The retina of the human eye displays according to \textit{Curcio et al.} \cite{Curcio1987} a particularly strong hexagonal arrangement of sensory cells, whereas following \textit{Middleton and Sivaswamy} \cite{Middleton2005} the processing of incoming signals is handled much more efficiently. Through the layers of the retina the reduction of the received information to the nerve fibers takes place, which are connected to the brain via afferents over the optic nerve. \textit{Hubel and Wiesel} \cite{Hubel1968} demonstrated that the image in the visual cortex is projected retinotopic, whereas adjacent structures of the retina are preserved and processed through the following areas of the brain.

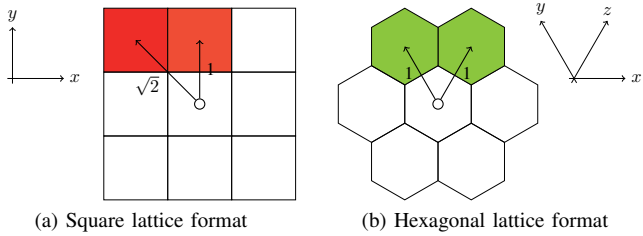
\begin{figure}[tb]
    \centering

    \subfloat[Square lattice format]{
        \scalebox{0.68}{\begin{tikzpicture}[r/.style={draw, rectangle, minimum size=12.5mm}]
            \node[r, fill=Red!95] (r1) at (-1.25,  1.25) {};
            \node[r, fill=Red!85] (r2) at ( 0,     1.25) {};
            \node[r]              (r3) at ( 1.25,  1.25) {};
            \node[r]              (r4) at (-1.25,  0)    {};
            \node[r]              (r5) at ( 0,     0)    {};
            \node[r]              (r6) at ( 1.25,  0)    {};
            \node[r]              (r7) at (-1.25, -1.25) {};
            \node[r]              (r8) at ( 0,    -1.25) {};
            \node[r]              (r9) at ( 1.25, -1.25) {};

            \node[draw, circle, minimum size=2mm, inner sep=0, semithick] (c) at (0, 0) {};

            \draw[->, semithick] (c) -- node[left, yshift=-3mm] {$\sqrt{2}$} (-1.25, 1.25);
            \draw[->, semithick] (c) -- node[right]             {1}          ( 0,    1.25);

            \draw[->, shorten <=-1mm] (-3.66, 0.5) -- (-2.66, 0.5) node[right] {$x$};
            \draw[->, shorten <=-1mm] (-3.66, 0.5) -- (-3.66, 1.5) node[above] {$y$};
        \end{tikzpicture}}}
    \quad
    \subfloat[Hexagonal lattice format]{
        \scalebox{0.68}{\begin{tikzpicture}[h/.style={draw, shape=regular polygon, regular polygon sides=6, rotate=30, minimum size=15mm}]
            \node[h]                 (0) at ( 0,        0)     {};
            \node[h]                 (1) at ( 1.3,      0)     {};
            \node[h, fill=LimeGreen] (2) at ( 1.3 / 2,  1.125) {};
            \node[h, fill=LimeGreen] (3) at (-1.3 / 2,  1.125) {};
            \node[h]                 (4) at (-1.3,      0)     {};
            \node[h]                 (5) at (-1.3 / 2, -1.125) {};
            \node[h]                 (6) at ( 1.3 / 2, -1.125) {};

            \node[draw, circle, minimum size=2mm, inner sep=0, semithick] (c) at (0, 0) {};

            \draw[->, semithick] (c) -- node[right] {1} ( 1.3 / 2, 1.125);
            \draw[->, semithick] (c) -- node[left]  {1} (-1.3 / 2, 1.125);

            \draw[->, shorten <=-1mm] (2.66, 0.5) -- (3.66,           0.5)         node[right] {$x$};
            \draw[->, shorten <=-1mm] (2.66, 0.5) -- (2.66 - 1.3 / 2, 0.5 + 1.125) node[above] {$y$};
            \draw[->, shorten <=-1mm] (2.66, 0.5) -- (2.66 + 1.3 / 2, 0.5 + 1.125) node[above] {$z$};
        \end{tikzpicture}}}

    \caption{Square and hexagonal lattice format in comparison.}
    \label{figure:lattice_formats}
\end{figure}

The use of hexagonal structures emerges therefore as an evolutionarily motivated approach. Compared to its square counterpart, the hexagonal lattice format (Fig.~\ref{figure:lattice_formats}) features a number of decisive advantages. These include the homogeneity of the hexagonal lattice format, whereby the equidistance and uniqueness of neighborhood as well as an increased radial symmetry are given. Together with an by $13.4$~\% increased transformation efficiency as shown by \textit{Petersen and Middleton} \cite{Petersen1962} and \textit{Mersereau} \cite{Mersereau1979}, hexagonal representations enable the storage of larger amounts of data on the same number of sampling points, which following \textit{Golay} \cite{Golay1969} result in reduced computation times, less quantization errors, and an increased efficiency in programming.
%%%%%%%%%%%%%%%%%%%%%%%%%%%%%%%%%%%%% Text %%%%%%%%%%%%%%%%%%%%%%%%%%%%%%%%%%%%%

\subsection{Related work}

%%%%%%%%%%%%%%%%%%%%%%%%%%%%%%%%%%%%% Text %%%%%%%%%%%%%%%%%%%%%%%%%%%%%%%%%%%%%
Current hexagonal image processing research includes applications in the fields of observation, experiment, and simulation in ecology by \textit{Birch et al.} \cite{Birch2007}, the design of geodesic grid systems by \textit{Sahr et al.} \cite{Sahr2003}, sensor-based image processing and remote sensing \cite{Hauschild1996, Jung1999, Ambrosio2001}, medical imaging \cite{Lin1996, Schwarz1999, Neeser2000}, and image synthesis \cite{Theussl2001}.

Furthermore, first experimental results for supply demand forecasting by \textit{Ke et al.} (2018) \cite{Ke2018} and the analysis of atmospheric telescope data for event detection and classification by \textit{Erdmann et al.} (2018) \cite{Erdmann2018} and \textit{Shilon et al.} (2019) \cite{Shilon2019} give an overview over currently emerging application areas for hexagonal image processing in the context of machine learning. These application areas show promising opportunities as current research often deploys an initial processing step to transform the given hexagonal lattice format based image data into a square representation \cite{Feng2016, Mangano2018}.

As the amount of research which combines principles from hexagonal image processing and machine learning is quite limited, this contribution also aims to overcome the downsides of currently developed hexagonal machine learning approaches. These include not only the development of hexagonal convolutional layers following \textit{Hoogeboom et al.} (2018) \cite{Hoogeboom2018} and \textit{Steppa and Holch} (2019) \cite{Steppa2019}, but also the implementation of the underlying addressing scheme as well as the necessary processing steps for transformation and visualization.
%%%%%%%%%%%%%%%%%%%%%%%%%%%%%%%%%%%%% Text %%%%%%%%%%%%%%%%%%%%%%%%%%%%%%%%%%%%%

\subsection{Contribution of this work}

%%%%%%%%%%%%%%%%%%%%%%%%%%%%%%%%%%%%% Text %%%%%%%%%%%%%%%%%%%%%%%%%%%%%%%%%%%%%
In order to combine the advantages of the research fields of hexagonal image processing and machine learning in the form of deep neural networks, this contribution serves as a general application-oriented approach the conception of a first hexagonal deep learning framework. While serving the recognition and classification of objects in larger data sets where the original image resolution often needs to be reduced as an initial processing step, the necessary information content should also be maximized. With focus on the implementation of the underlying structure, the designed framework, called Hexnet, includes the necessary processing steps as well as its underlying hexagonal addressing scheme.

However, as hexagonal image data has to be captured using a hexagonal sensor, and the corresponding hardware is either rare or limited in its applicability, alternatives include the transformation or synthesis of hexagonal image data. Therefore, the question arises, in which way and to what degree currently deployed square lattice format based data sets can be leveraged in the context of hexagonal deep learning approaches.

As this contribution aims to benefit both researchers and users alike by providing the necessary tools to ease the application and development of hexagonal models, we will also provide the tools to evaluate said models with focus on their performance and transformation quality as compared to classical square and hexagonal lattice format based approaches. The related developed architectures, procedures, and test results, including future developments and further application-specific contributions, are made publicly available to the research community via open science and are found on the project page of Hexnet and its repository\footnote{\url{https://github.com/TSchlosser13/Hexnet}}.
%%%%%%%%%%%%%%%%%%%%%%%%%%%%%%%%%%%%% Text %%%%%%%%%%%%%%%%%%%%%%%%%%%%%%%%%%%%%

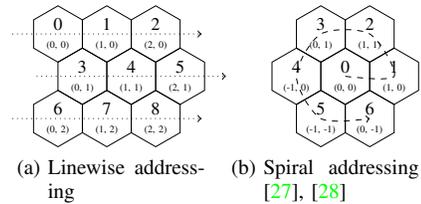
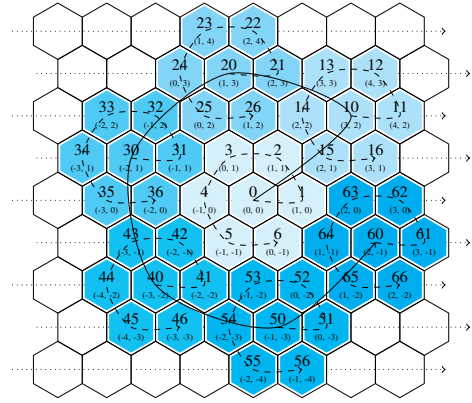
\begin{figure}[tb]
    \centering

    \subfloat[Linewise addressing]{
        \scalebox{0.75}{\begin{tikzpicture}[
         h/.style={shape=regular polygon, regular polygon sides=6, shape border rotate=30, minimum size=10mm},
         align=center, font=\small, inner sep=0]
            \drawhexarray{0}{draw}{1}{1}{2}{2}
        \end{tikzpicture}}}
    \quad
    \subfloat[Spiral addressing \cite{Burt1980, Gibson1982}]{
        \scalebox{0.75}{\begin{tikzpicture}[
         h/.style={shape=regular polygon, regular polygon sides=6, shape border rotate=30, minimum size=10mm},
         align=center, font=\small, inner sep=0]
            \drawhexarray{0}{draw=none}{0}{0}{2}{2}
            \drawhexarraysaaoffirstorder{0}{draw}{1}{1}{0}{0}{0}{}
        \end{tikzpicture}}}

    \subfloat[Combination of (a) and (b)]{
        \scalebox{0.75}{\begin{tikzpicture}[
         h/.style={shape=regular polygon, regular polygon sides=6, shape border rotate=30, minimum size=10mm},
         align=center, font=\small, inner sep=0]
            \drawhexarray{0}{draw}{0}{1}{7}{8}
            \drawhexarraysaaofsecondorder{0}{draw=none, scale=0.66}{4}{4}{0}{0}{0}{cyan}
        \end{tikzpicture}}}

    \caption{Construction of hexagonal addressing schemes.}
    \label{figure:addressing_schemes}
\end{figure}

\begin{figure*}[tb]
    \centering

        \begin{tikzpicture}[
         h/.style={shape=regular polygon, regular polygon sides=6, shape border rotate=30, minimum size=10mm},
         align=center, font=\small, inner sep=0]
            \node (input) at (0, 0) {\includegraphics[height=3cm]{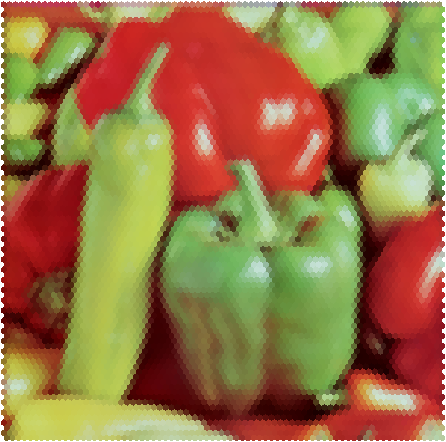}};
            \node[draw, rectangle, minimum size=2mm] (input_roi) at (input) {};

            \begin{scope}[xshift=3.5cm, scale=0.33, transform shape]
                \foreach \i in {0, ..., 5} {
                    \node[draw, rectangle, fill=white, minimum size=3cm] (l0_ha\i) at (0.33 * \i, -0.33 * \i) {};
                }

                \node[draw, rectangle, minimum size=6mm] (l0_ha5_roi) at (l0_ha5) {};
            \end{scope}

            \begin{scope}[xshift=6.5cm, scale=0.33, transform shape]
                \foreach \i in {0, ..., 5} {
                    \ha{l1_ha\i}{draw, fill=white, shift={(3.3 * \i mm, -3.3 * \i mm)}}{0}{0}{7}{8}
                }

                \node[h, fill=Red!95,             minimum size=8mm] at (l1_ha5_h_3_3) {};
                \node[h, fill=Red!85,             minimum size=8mm] at (l1_ha5_h_4_3) {};
                \node[h, fill=LimeGreen,          minimum size=8mm] at (l1_ha5_h_3_4) {};
                \node[h, fill=LimeGreen!50!black, minimum size=8mm] at (l1_ha5_h_4_4) {};
                \node[h, fill=LimeGreen,          minimum size=8mm] at (l1_ha5_h_5_4) {};
                \node[h, fill=LimeGreen,          minimum size=8mm] at (l1_ha5_h_3_5) {};
                \node[h, fill=LimeGreen,          minimum size=8mm] at (l1_ha5_h_4_5) {};
            \end{scope}

            \begin{scope}[xshift=9.5cm, scale=0.33, transform shape]
                \foreach \i in {0, ..., 5} {
                    \ha{l2_ha\i}{draw, fill=white, shift={(3.3 * \i mm, -3.3 * \i mm)}}{0}{0}{2}{2}
                }

                \node[h, fill=LimeGreen, minimum size=8mm] at (l2_ha5_h_1_1) {};
            \end{scope}

            \begin{scope}[xshift=12.5cm, scale=0.33, transform shape]
                \newcommand*{\cvdots}{\vcenter{\hbox{.}\hbox{.}\hbox{.}}}

                \node[draw, h] (l3_h1) at (0,  3) {};
                \node          (l3_hd) at (0,  0) {\Huge\bfseries$\cvdots$};
                \node[draw, h] (l3_h2) at (0, -3) {};

                \node[draw, h] (l4_h1) at (3,  3) {};
                \node          (l4_hd) at (3,  0) {\Huge\bfseries$\cvdots$};
                \node[draw, h] (l4_h2) at (3, -3) {};

                \draw[->] (l3_h1) -- (l4_h1);
                \draw[->] (l3_h1) -- (l4_h2);
                \draw[->] (l3_h2) -- (l4_h1);
                \draw[->] (l3_h2) -- (l4_h2);
            \end{scope}

            \draw (input_roi.north east) -- (l0_ha5.center);
            \draw (input_roi.south east) -- node[below=15mm] {\textit{CL\textsubscript{1}} ($7^1$),\nl \textit{PL\textsubscript{1}} ($7^1$)} (l0_ha5.center);

            \draw (l0_ha5_roi.north east) -- (l1_ha5_h_4_4.center);
            \draw (l0_ha5_roi.south east) -- node[below=15mm] {\textit{CL\textsubscript{2}} ($7^1$),\nl \textit{PL\textsubscript{2}} ($7^1$)} (l1_ha5_h_4_4.center);

            \draw (l1_ha5_h_4_3.north east) -- (l2_ha5_h_1_1.center);
            \draw (l1_ha5_h_4_5.south east) -- node[below=15mm] {\textit{CL\textsubscript{3}} ($7^1$),\nl \textit{PL\textsubscript{3}} ($7^1$)} (l2_ha5_h_1_1.center);

            \draw (l2_ha0_h_2_0.north east) -- (l3_h1);
            \draw (l2_ha5_h_2_2.south east) -- (l3_h2);

            \node[above=2mm of input]        {Input ($75 \times 86$)};
            \node[above=2mm of l0_ha0]       {\textit{FM\textsubscript{1}} ($6~@~25 \times 28$)};
            \node[above=2mm of l1_ha0_h_3_0] {\textit{FM\textsubscript{2}} ($6~@~8 \times 9$)};
            \node[above=4mm of l2_ha0_h_1_0] {\textit{FM\textsubscript{3}} ($6~@~3 \times 3$)};
            \node[above=4mm of l3_h1] at ($(l3_h1)!0.5!(l4_h1)$) {\textit{FCLs} + Output};
        \end{tikzpicture}

    \caption{Exemplary structure of a hexagonal convolutional neural network with convolutional (\textit{CL}) and pooling layers (\textit{PL}) of order $1$, fully connected layers (\textit{FCL}), and feature maps (\textit{FM}).}
    \label{figure:H-CNN}
\end{figure*}
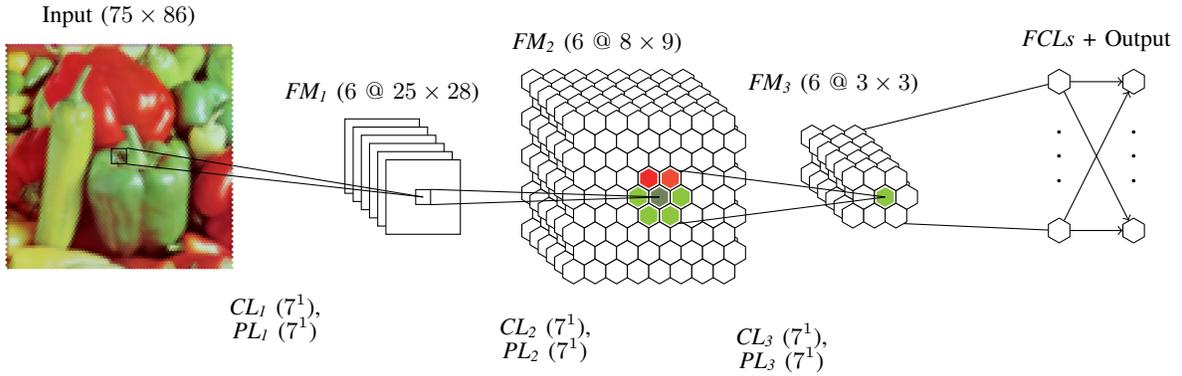

\section{Fundamentals and methods}

%%%%%%%%%%%%%%%%%%%%%%%%%%%%%%%%%%%%% Text %%%%%%%%%%%%%%%%%%%%%%%%%%%%%%%%%%%%%
To allow the addressing, transformation, and visualization of images in their hexagonal representation, the necessary processing steps as well as their underlying addressing scheme have to be determined. The following sections introduce our proposed hexagonal addressing scheme as identified for its applicability in comparison to currently deployed square and hexagonal lattice format based architectures. In conjunction with the proposed architecture, we will then introduce the principles of hexagonal image processing in the context of deep learning.
%%%%%%%%%%%%%%%%%%%%%%%%%%%%%%%%%%%%% Text %%%%%%%%%%%%%%%%%%%%%%%%%%%%%%%%%%%%%

\subsection{The hexagonal addressing scheme}

%%%%%%%%%%%%%%%%%%%%%%%%%%%%%%%%%%%%% Text %%%%%%%%%%%%%%%%%%%%%%%%%%%%%%%%%%%%%
While the one-dimensional spiral architecture addressing scheme (SAA) introduced by \textit{Burt} \cite{Burt1980} and \textit{Gibson and Lucas} \cite{Gibson1982} established itself as the preferred addressing scheme for hexagonal lattice format based image processing systems (Fig.~\ref{figure:addressing_schemes}b), it also reduces the locality and increases the complexity when transforming square lattice format based images into their hexagonal representation. Pseudohexagonal addressing schemes, however, are easier to implement, as every second row is shifted to the right \cite{Fitz1996}. \textit{Hoogeboom et al.} \cite{Hoogeboom2018} deploy an addressing scheme which is solely based on a skewed axis architecture. As this addressing scheme is easy to implement, it also greatly increases the amount of stored pixel data at its corners.

This contribution proposes a hybrid approach that combines the one-dimensional linewise architecture (Fig.~\ref{figure:addressing_schemes}a), which is common for square lattice format based addressing schemes, with SAA. By combining both addressing schemes, a given image can still be recovered as its original resolution is preserved. The resulting hierarchical hexagonal structures can furthermore be exploited based on the properties of their pyramidal decompositions, also called septrees, whereas a pseudohexagonal addressing scheme is implemented efficiently by storing the shifted Cartesian coordinates for every second row \cite{Her1994, Her1995}.

The one-dimensional spiral architecture is then constructed alongside the numbering shown in Fig.~\ref{figure:addressing_schemes}c, where a hexagonal pixel is denoted by a Hexint and a set of $7^n$ Hexints is referred to as a Hexarray of order $n$ ($n \in \mathbb{N}_{\geq 0}$) \cite{Middleton2005}. As based on SAA, the construction of each Hexarray of order $n$ consists of seven sub-Hexarrays of order $n - 1$. The image data itself, however, is stored based on the linewise addressing scheme, and the offsets for each sub-Hexarray are kept depending on the current Hexarray's digits, which assign each Hexint to a unique position inside the Hexarray. For transformation, the Hexarray is centered over the given image and interpolated using either nearest-neighbor, bilinear, or bicubic interpolation \cite{Her1994}. However, more complex approaches, including hexagonal splines \cite{VanDeVille2004}, are surely existing.

To allow the processing of Hexarrays other than given by SAA, a sub-Hexarray of order $1$ (Fig.~\ref{figure:addressing_schemes}b) can be extended by expanding its side length. The in \eqref{equation:H_N} shown, on ($x, y$) (Fig.~\ref{figure:lattice_formats}b) based hexagonal block $H_N$ with side length $N$ will be hereinafter deployed to support additional kernel sizes. The hexagonal block $H_2$ constitutes therefore an SAA-based kernel of order $1$ with size $M$ Hexints in \eqref{equation:M}.
%%%%%%%%%%%%%%%%%%%%%%%%%%%%%%%%%%%%% Text %%%%%%%%%%%%%%%%%%%%%%%%%%%%%%%%%%%%%

\begin{equation}
    \begin{split}
        H_N = \{ (x, y) : 0 < x < 2 \cdot N - 1, \\
        0 < y < 2 \cdot N - 1, |x - y| < N \}
    \end{split}
    \label{equation:H_N}
\end{equation}

\begin{equation}
    M = 3 \cdot N^2 - 3 \cdot N + 1
    \label{equation:M}
\end{equation}

\subsection{Conventional and hexagonal convolutional neural networks}

%%%%%%%%%%%%%%%%%%%%%%%%%%%%%%%%%%%%% Text %%%%%%%%%%%%%%%%%%%%%%%%%%%%%%%%%%%%%
Conventional convolutional neural networks, which are subdivided almost exclusively based on square lattice formats, hereinafter referred to as square CNNs, also called S-CNN, are as proposed by \textit{LeCun et al.} \cite{LeCun1998} subdivided into a set of alternating layers, consisting of convolutional (\textit{CL}), pooling (\textit{PL}), and fully connected layers (\textit{FCL}). Convolutional and pooling layer realize the filter-based convolution of features, abstract in their spatial dimensionality, and represent the extracted features in the form of feature maps (\textit{FM}) according to their current depth in the network. The following fully connected layers are used to classify the represented features. Finally, in order to be able to find an objective function of a given optimization problem, e.g. for object recognition and classification, a backpropagation based on the provided data for training takes place, followed by an update of the network and its parameters.

The hexagonal convolutional neural network shown in Fig.~\ref{figure:H-CNN}, also called H-CNN, is based on the addressing scheme for hexagonal architectures introduced in Fig.~\ref{figure:addressing_schemes}c and demonstrates the functionality using the example of the test image \textit{Pepper} \cite{Weber1993}. The results shown in each layer represent for \textit{FM\textsubscript{1}} to \textit{FM\textsubscript{3}} the feature maps of the respective convolution with subsequent pooling, whereby the filters are given by the sub-Hexarrays of the proposed hexagonal addressing scheme.
%%%%%%%%%%%%%%%%%%%%%%%%%%%%%%%%%%%%% Text %%%%%%%%%%%%%%%%%%%%%%%%%%%%%%%%%%%%%

\begin{figure*}[tb]
    \centering

    \includegraphics[width=0.07\textwidth, align=c]{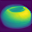}
    \includegraphics[width=0.07\textwidth, align=c]{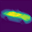}
    \includegraphics[width=0.07\textwidth, align=c]{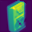}
    \includegraphics[width=0.07\textwidth, align=c]{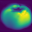}
    \includegraphics[width=0.07\textwidth, align=c]{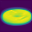}
    \includegraphics[width=0.07\textwidth, align=c]{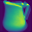}
    \includegraphics[width=0.07\textwidth, align=c]{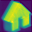}
    \includegraphics[width=0.07\textwidth, align=c]{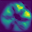}
    \includegraphics[width=0.07\textwidth, align=c]{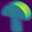}
    \includegraphics[width=0.07\textwidth, align=c]{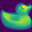}
    \vspace{0.2cm}

    \includegraphics[width=0.07\textwidth, align=c]{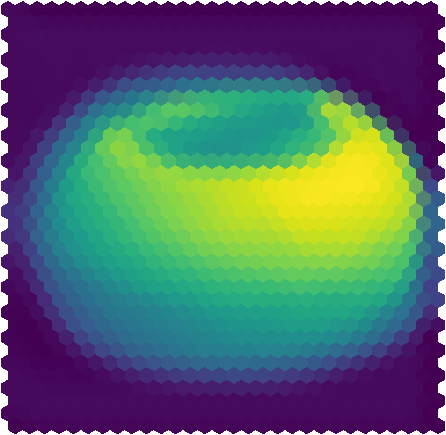}
    \includegraphics[width=0.07\textwidth, align=c]{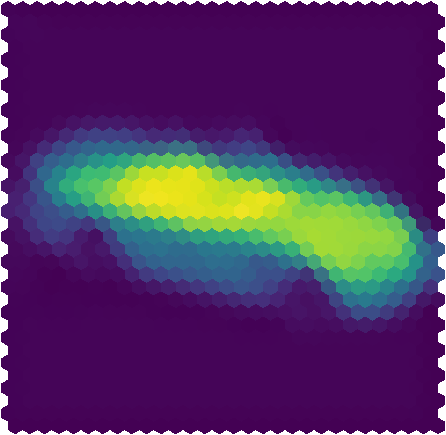}
    \includegraphics[width=0.07\textwidth, align=c]{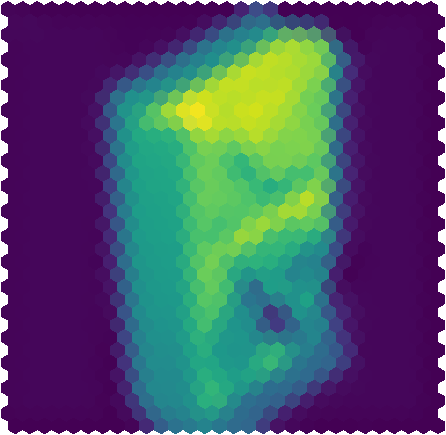}
    \includegraphics[width=0.07\textwidth, align=c]{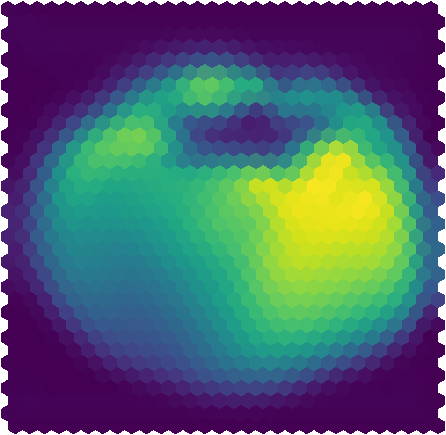}
    \includegraphics[width=0.07\textwidth, align=c]{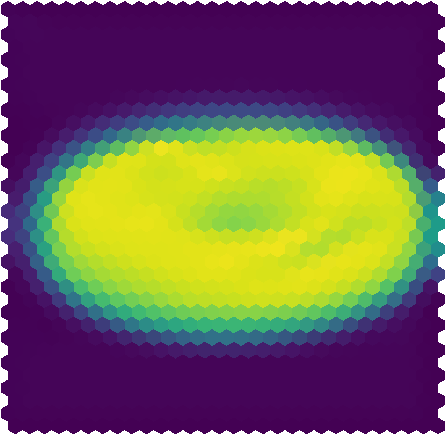}
    \includegraphics[width=0.07\textwidth, align=c]{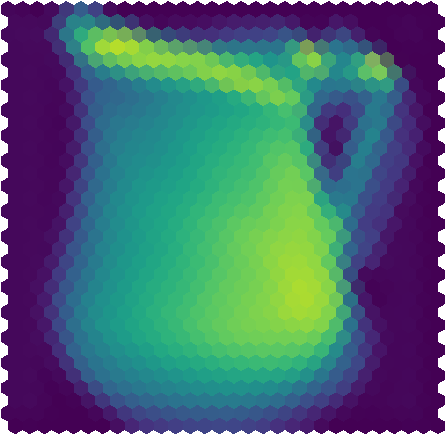}
    \includegraphics[width=0.07\textwidth, align=c]{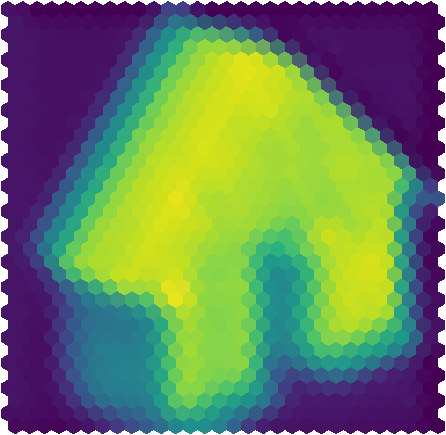}
    \includegraphics[width=0.07\textwidth, align=c]{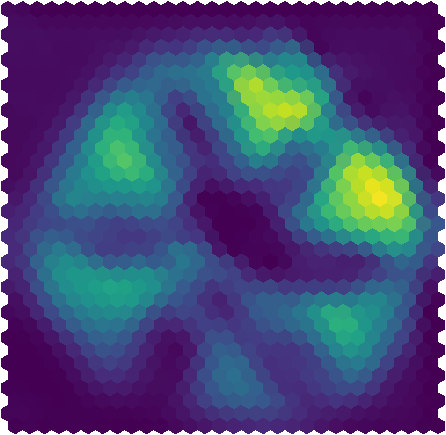}
    \includegraphics[width=0.07\textwidth, align=c]{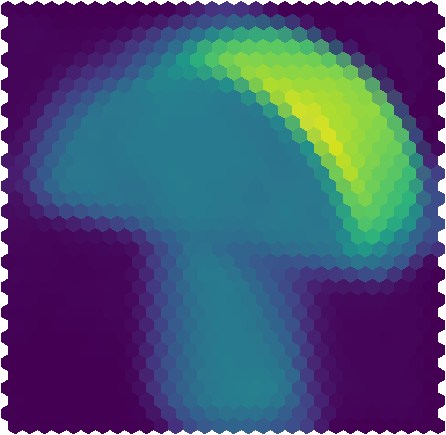}
    \includegraphics[width=0.07\textwidth, align=c]{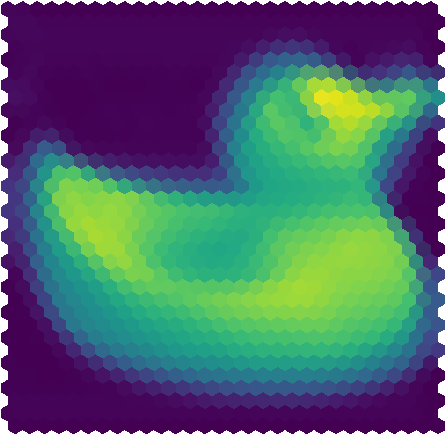}
    \vspace{0.2cm}

    \caption{Visualized feature maps for exemplary S-CNN (t.) and H-CNN (b.) test results with one convolutional layer. Shown are ten different classes for the COIL-100 data set \cite{Nene1996b}.}
    \label{figure:feature_maps}
\end{figure*}

\begin{figure}[tb]
    \centering

    \subfloat[Even rows]{
        \scalebox{0.75}{\begin{tikzpicture}[
         h/.style={shape=regular polygon, regular polygon sides=6, shape border rotate=30, minimum size=10mm},
         align=center, font=\small, inner sep=0]
            \drawhexarray{0}{draw}{0}{0}{2}{3}
            \drawhexarraysaaoffirstorder{0}{draw=none, scale=0.66}{1}{2}{0}{0}{0}{cyan}
        \end{tikzpicture}}}
    \quad
    \subfloat[Odd rows]{
        \scalebox{0.75}{\begin{tikzpicture}[
         h/.style={shape=regular polygon, regular polygon sides=6, shape border rotate=30, minimum size=10mm},
         align=center, font=\small, inner sep=0]
            \drawhexarray{0}{draw}{0}{0}{2}{3}
            \drawhexarraysaaoffirstorder{0}{draw=none, scale=0.66}{1}{1}{0}{0}{0}{cyan}
        \end{tikzpicture}}}

    \caption{Hexagonal convolutional and pooling layer kernels for even and odd row indices as based on their underlying addressing scheme.}
    \label{figure:kernels_even_odd_rows}
\end{figure}
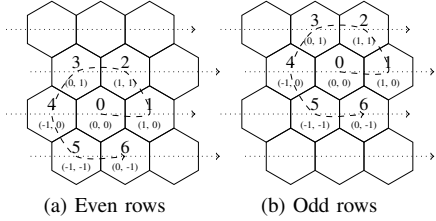

%%%%%%%%%%%%%%%%%%%%%%%%%%%%%%%%%%%%% Text %%%%%%%%%%%%%%%%%%%%%%%%%%%%%%%%%%%%%
Translational equivariance as well as translational and scaling invariance are according to \textit{Cohen and Welling} \cite{Cohen2016} decisive features for CNNs in square lattice based formats, while following \textit{Hoogeboom et al.} \cite{Hoogeboom2018} hexagonal representations provide the advantages of rotational invariance due to the substructures of different orders. In particular, following \textit{Dieleman et al.} \cite{Dieleman2016}, the need for data augmentation can be significantly reduced. Together with the increased axial symmetry of hexagonal representations, group convolutions can be deployed, which furthermore show a reduced angular dependence and result in increased training and test rates \cite{Hoogeboom2018, Cohen2016, Dieleman2016}.
%%%%%%%%%%%%%%%%%%%%%%%%%%%%%%%%%%%%% Text %%%%%%%%%%%%%%%%%%%%%%%%%%%%%%%%%%%%%

\section{Implementation of the hexagonal deep learning Framework}

%%%%%%%%%%%%%%%%%%%%%%%%%%%%%%%%%%%%% Text %%%%%%%%%%%%%%%%%%%%%%%%%%%%%%%%%%%%%
To enable the application of hexagonal machine learning in the form of hexagonal deep neural networks as well as to ease their use and integration, including hexagonal models and layers into already existing ones, without any additional overhead, we based our framework on the currently most commonly in research and application deployed machine learning framework TensorFlow\footnote{\url{https://www.tensorflow.org}, version 2.0.0-beta1} with Keras\footnote{\url{https://keras.io}} as its front end.
%%%%%%%%%%%%%%%%%%%%%%%%%%%%%%%%%%%%% Text %%%%%%%%%%%%%%%%%%%%%%%%%%%%%%%%%%%%%

\subsection{Hexagonal layers}

%%%%%%%%%%%%%%%%%%%%%%%%%%%%%%%%%%%%% Text %%%%%%%%%%%%%%%%%%%%%%%%%%%%%%%%%%%%%
Hexagonal layers for H-CNNs encompass as compared to conventional square lattice format based ones hexagonal convolutional layers as well as pooling layers. Fully connected layers, however, are applied to either square or hexagonal lattice format based inputs, as a mapping between two adjacent layers is established for every given neuron.
%%%%%%%%%%%%%%%%%%%%%%%%%%%%%%%%%%%%% Text %%%%%%%%%%%%%%%%%%%%%%%%%%%%%%%%%%%%%

\subsubsection{Convolutional layer}

%%%%%%%%%%%%%%%%%%%%%%%%%%%%%%%%%%%%% Text %%%%%%%%%%%%%%%%%%%%%%%%%%%%%%%%%%%%%
Current research by \textit{Steppa and Holch} \cite{Steppa2019} suggest a hexagonal convolutional layer implementation where each hexagonal block of pixels is separated by a number of $\lfloor\sfrac{d}{2}\rfloor$ vertical slices with $d$ denoting the horizontal diameter of the hexagonal pixel block. These hexagonal sub-blocks are then convolved using $d$ convolutions with their respective kernels.

As the number of slices and therefore convolutions increases for kernels of higher side length, and the additional convolutions prove to be computational costly, the number of required convolutions should be kept as small as possible. The number of necessary convolutions is reduced to two by differentiating between kernels for even and for odd row indices. Figure~\ref{figure:kernels_even_odd_rows} shows this by means of two kernels with a side length of $2$, one for each row type as based on their underlying addressing scheme. This step is important, as shifted rows also begin with a starting index of $0$.

To select a set of hexagonal blocks, the hexagonal convolutional layer is facilitated by masking the input image batch to obtain the current regions of interest. Therefore, two different masking matrices have to be considered, one for each row type, whereas row indices are identified by the position of the leftmost Hexint of the current row. The feature maps depicted in Fig.~\ref{figure:feature_maps} show exemplary S-CNN (top) and H-CNN (bottom) test results with one convolutional layer for the COIL-100 data set \cite{Nene1996b}.
%%%%%%%%%%%%%%%%%%%%%%%%%%%%%%%%%%%%% Text %%%%%%%%%%%%%%%%%%%%%%%%%%%%%%%%%%%%%

\subsubsection{Pooling layer}

%%%%%%%%%%%%%%%%%%%%%%%%%%%%%%%%%%%%% Text %%%%%%%%%%%%%%%%%%%%%%%%%%%%%%%%%%%%%
The pooling layer, as inspired by the scaling of Hexarrays by \textit{Sheridan et al.} \cite{Sheridan2000} and \textit{Middleton and Sivaswamy} \cite{Middleton2005}, is implemented based on the sub-Hexarray orders of the realized addressing scheme. Square and hexagonal pooling layers differ by the fact that kernels based on classical square lattice formats are often in square size and hexagonal ones show different offsets corresponding to the deployed pooling size.

To solve the mapping problem of the resulting hexagonal pooling kernel offsets with or without strides corresponding to the current sub-Hexarray offset mapping, the offsets have to be considered in a more general manner, including strides that are not given by the SAA-based offsets of different order. The resulting linear assignment problem in \eqref{equation:G} with offset sets $O_n$ and $O_{n - 1}$ for the Hexarray orders $n$ and $n - 1$ of the respective pooling kernels constitutes the basis of this problem. The general assignment problem is then solved by deploying linear programming or, more specifically, the Hungarian method \cite{Kuhn1955} by minimizing its $\ell 2$-norm based cost.
%%%%%%%%%%%%%%%%%%%%%%%%%%%%%%%%%%%%% Text %%%%%%%%%%%%%%%%%%%%%%%%%%%%%%%%%%%%%

\begin{equation}
    G = (O_n \cup O_{n - 1}, E), |O_n| = |O_{n - 1}|
    \label{equation:G}
\end{equation}

%%%%%%%%%%%%%%%%%%%%%%%%%%%%%%%%%%%%% Text %%%%%%%%%%%%%%%%%%%%%%%%%%%%%%%%%%%%%
The resulting assignment also results in a slight shift and rotation \cite{Sheridan2000, Middleton2005}, as Hexarrays of order $n$ can not be mapped onto sub-Hexarrays of order $n - 1$ without introducing additional invariance.
%%%%%%%%%%%%%%%%%%%%%%%%%%%%%%%%%%%%% Text %%%%%%%%%%%%%%%%%%%%%%%%%%%%%%%%%%%%%

\subsection{Visualization}

%%%%%%%%%%%%%%%%%%%%%%%%%%%%%%%%%%%%% Text %%%%%%%%%%%%%%%%%%%%%%%%%%%%%%%%%%%%%
In order to ensure an adequate visualization of the hexagonal lattice format by Hexnet, a general-purpose graphics processing units (GPGPU) accelerated representation was realized. Thus, according to the current level of detail, a shader-based approximation of each Hexint takes place.
%%%%%%%%%%%%%%%%%%%%%%%%%%%%%%%%%%%%% Text %%%%%%%%%%%%%%%%%%%%%%%%%%%%%%%%%%%%%

\begin{table}[tb]
    \renewcommand{\arraystretch}{1.2}
    \centering

    \begin{tabular}{|M{1cm}|M{1.9cm}|M{1cm}|M{1.6cm}|M{1cm}|}
        \hline
        Layer & Type & Size / rate & Output shape & Param \# \\
        \hline
        \noalign{\vskip 2pt}

        \hline
        conv1    & SConv2D    & $3 \times 3$ & $(32, 32, 32)$ & $896$ \\
        \hline
        conv2    & SConv2D    & $3 \times 3$ & $(16, 16, 32)$ & $9\thinspace 248$ \\
        \hline
        dropout1 & Dropout    & $0.25$       & $(16, 16, 32)$ & $0$ \\
        \hline
        conv3    & SConv2D    & $3 \times 3$ & $(16, 16, 64)$ & $18\thinspace 496$ \\
        \hline
        pool     & SMaxPool2D & $2 \times 2$ & $(8, 8, 64)$   & $0$ \\
        \hline
        dropout2 & Dropout    & $0.25$       & $(8, 8, 64)$   & $0$ \\
        \hline
        flatten  & Flatten    & /            & $(4096)$       & $0$ \\
        \hline
        dense1   & Dense      & $128$        & $(128)$        & $524\thinspace 416$ \\
        \hline
        dropout3 & Dropout    & $0.5$        & $(128)$        & $0$ \\
        \hline
        dense2   & Dense      & $100$        & $(100)$        & $12\thinspace 900$ \\
        \hline
        \noalign{\vskip 2pt}

        \cline{1-3}
        \multicolumn{3}{|c|}{Total trainable params: $565\thinspace 956$} & \multicolumn{2}{c}{} \\
        \cline{1-3}
    \end{tabular}

    \caption{S-CNN layer configuration, consisting of two convolutional blocks and one fully connected layer.}
    \label{table:S-CNN_model_configuration}
\end{table}

\begin{table}[tb]
    \renewcommand{\arraystretch}{1.2}
    \centering

    \begin{tabular}{|M{1cm}|M{1.9cm}|M{1cm}|M{1.6cm}|M{1cm}|}
        \hline
        Layer & Type & Size / rate & Output shape & Param \# \\
        \hline
        \noalign{\vskip 2pt}

        \hline
        conv1    & HConv2D    & $7^1$  & $(34, 30, 32)$ & $704$ \\
        \hline
        conv2    & HConv2D    & $7^1$  & $(18, 15, 32)$ & $7\thinspace 200$ \\
        \hline
        dropout1 & Dropout    & $0.25$ & $(18, 15, 32)$ & $0$ \\
        \hline
        conv3    & HConv2D    & $7^1$  & $(18, 15, 64)$ & $14\thinspace 400$ \\
        \hline
        pool     & HMaxPool2D & $7^1$  & $(8, 5, 64)$   & $0$ \\
        \hline
        dropout2 & Dropout    & $0.25$ & $(8, 5, 64)$   & $0$ \\
        \hline
        flatten  & Flatten    & /      & $(2560)$       & $0$ \\
        \hline
        dense1   & Dense      & $128$  & $(128)$        & $327\thinspace 808$ \\
        \hline
        dropout3 & Dropout    & $0.5$  & $(128)$        & $0$ \\
        \hline
        dense2   & Dense      & $100$  & $(100)$        & $12\thinspace 900$ \\
        \hline
        \noalign{\vskip 2pt}

        \cline{1-3}
        \multicolumn{3}{|c|}{Total trainable params: $363\thinspace 012$} & \multicolumn{2}{c}{} \\
        \cline{1-3}
    \end{tabular}

    \caption{H-CNN layer configuration, consisting of two convolutional blocks and one fully connected layer.}
    \label{table:H-CNN_model_configuration}
\end{table}

\section{Experimental test results, evaluation, and discussion}

%%%%%%%%%%%%%%%%%%%%%%%%%%%%%%%%%%%%% Text %%%%%%%%%%%%%%%%%%%%%%%%%%%%%%%%%%%%%
As current research that combines principles from the domain of hexagonal image processing with machine learning often only provides application-specific test results with data which were either captured using a hexagonal sensor or synthesized, this contribution also aims to provide a more general evaluation. This includes not only a comparison between conventional square lattice based DNNs and hexagonal ones, but also an evaluation regarding the hexagonal image transformation, which is deployed as a necessary initial processing step. Therefore, the transformation quality for both lattice formats has to be determined. Whereas S-DNNs and H-DNNs are evaluated by the number of used training parameters and resulting train and test rates, we will also consider the performance of our proposed addressing scheme in the context of alternative hexagonal image processing frameworks.
%%%%%%%%%%%%%%%%%%%%%%%%%%%%%%%%%%%%% Text %%%%%%%%%%%%%%%%%%%%%%%%%%%%%%%%%%%%%

\begin{figure}[tb]
    \centering

    \subfloat[]{
        \begin{tikzpicture}
            \begin{axis}[
             cycle list/Set1,
             legend style={anchor=north, at={(0.5, -0.25)}, legend columns=3},
             width=0.5\textwidth, height=0.33\textwidth, xlabel={Epoch}, ylabel={Accuracy},
              yticklabel style={/pgf/number format/.cd, fixed, fixed zerofill, precision=1}]
                \addplot+[restrict x to domain=0:25] table[x expr=\coordindex, y index=0] {data/CINIC-10_3x3.csv};
                \addlegendentry{CINIC\textsubscript{$3 \times 3$}}
                \addplot+[restrict x to domain=0:25] table[x expr=\coordindex, y index=0] {data/CINIC-10_2x2.csv};
                \addlegendentry{CINIC\textsubscript{$2 \times 2$}}
                \addplot+[restrict x to domain=0:25] table[x expr=\coordindex, y index=0] {data/CINIC-10_7.csv};
                \addlegendentry{CINIC\textsubscript{$7^1$}}

                \addplot+[restrict x to domain=0:25] table[x expr=\coordindex, y index=0] {data/COIL-100_3x3.csv};
                \addlegendentry{COIL\textsubscript{$3 \times 3$}}
                \addplot+[restrict x to domain=0:25] table[x expr=\coordindex, y index=0] {data/COIL-100_2x2.csv};
                \addlegendentry{COIL\textsubscript{$2 \times 2$}}
                \addplot+[restrict x to domain=0:25] table[x expr=\coordindex, y index=0] {data/COIL-100_7.csv};
                \addlegendentry{COIL\textsubscript{$7^1$}}

                \addplot+[restrict x to domain=0:25] table[x expr=\coordindex, y index=0] {data/MNIST_3x3.csv};
                \addlegendentry{MNIST\textsubscript{$3 \times 3$}}
                \addplot+[restrict x to domain=0:25] table[x expr=\coordindex, y index=0] {data/MNIST_2x2.csv};
                \addlegendentry{MNIST\textsubscript{$2 \times 2$}}
                \addplot+[restrict x to domain=0:25] table[x expr=\coordindex, y index=0] {data/MNIST_7.csv};
                \addlegendentry{MNIST\textsubscript{$7^1$}}
            \end{axis}
        \end{tikzpicture}}

    \subfloat[]{
        \begin{tikzpicture}
            \begin{axis}[
             cycle list/Set1,
             legend style={anchor=north, at={(0.5, -0.25)}, legend columns=3},
             width=0.5\textwidth, height=0.33\textwidth, xlabel={Epoch}, ylabel={Accuracy},
              yticklabel style={/pgf/number format/.cd, fixed, fixed zerofill, precision=1}]
                \addplot table[x expr=\coordindex, y index=0] {data/COIL-100_3x3.csv};
                \addlegendentry{COIL\textsubscript{$3 \times 3$}}
                \addplot table[x expr=\coordindex, y index=0] {data/COIL-100_2x2.csv};
                \addlegendentry{COIL\textsubscript{$2 \times 2$}}
                \addplot table[x expr=\coordindex, y index=0] {data/COIL-100_7.csv};
                \addlegendentry{COIL\textsubscript{$7^1$}}

                \addplot table[x expr=\coordindex, y index=0] {data/D-COIL-100_3x3.csv};
                \addlegendentry{D-COIL\textsubscript{$3 \times 3$}}
                \addplot table[x expr=\coordindex, y index=0] {data/D-COIL-100_2x2.csv};
                \addlegendentry{D-COIL\textsubscript{$2 \times 2$}}
                \addplot table[x expr=\coordindex, y index=0] {data/D-COIL-100_7.csv};
                \addlegendentry{D-COIL\textsubscript{$7^1$}}
            \end{axis}
        \end{tikzpicture}}

    \caption{S-CNN and H-CNN training results for the CINIC-10 \cite{Darlow2018}, COIL-100 \cite{Nene1996b}, and MNIST \cite{LeCun1998} data sets. Shown are the training results over 25 epochs for the transformed versions of the data sets (a) as well as COIL-100 with its transformed and additionally distorted version over 100 epochs (b), each averaged over five runs.}
    \label{figure:CNN_test_results}
\end{figure}
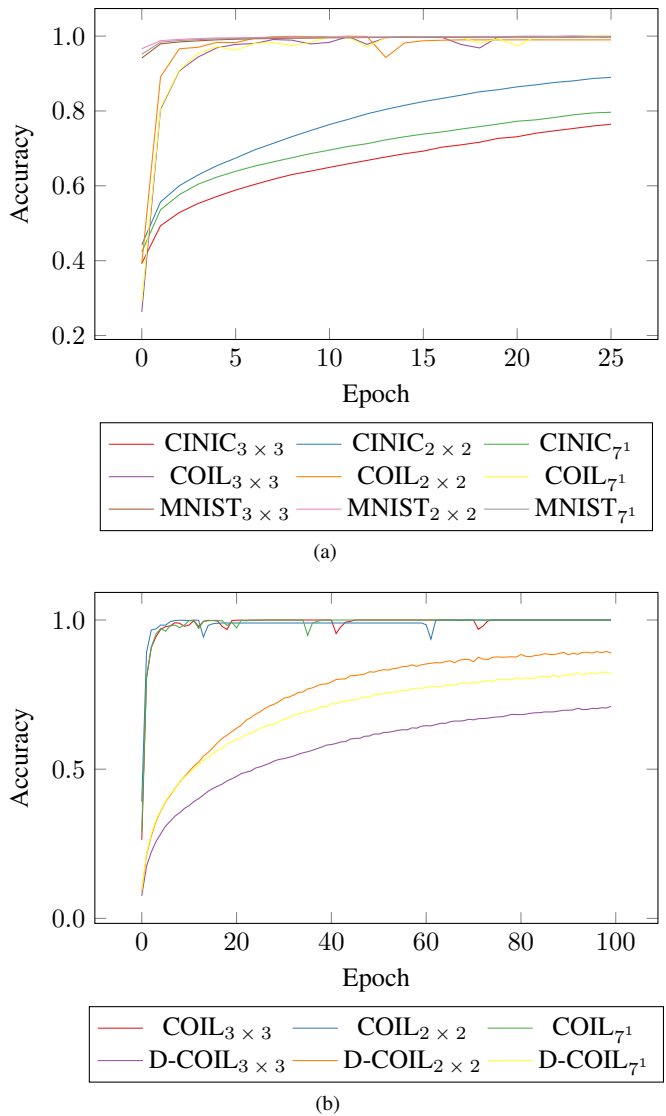

\begin{table}[tb]
    \renewcommand{\arraystretch}{1.2}
    \centering

    \begin{tabular}{|M{2cm}|M{2cm}|M{2cm}|}
        \hline
        Data set & Test run & Mean accuracy \\
        \hline
        \noalign{\vskip 2pt}

        \hline
        \multirow{3}{*}{CINIC-10}   & $3 \times 3$ & $0.809 \pm 0.02$ \\
        \cline{2-3}
                                    & $2 \times 2$ & $\mathbf{0.859 \pm 0.02}$ \\
        \cline{2-3}
                                    & $7^1$        & $0.831 \pm 0.02$ \\
        \hline
        \multirow{3}{*}{COIL-100}   & $3 \times 3$ & $0.987 \pm 0.01$ \\
        \cline{2-3}
                                    & $2 \times 2$ & $\mathbf{0.993 \pm 0.01}$ \\
        \cline{2-3}
                                    & $7^1$        & $0.989 \pm 0.01$ \\
        \hline
        \multirow{3}{*}{D-COIL-100} & $3 \times 3$ & $0.724 \pm 0.03$ \\
        \cline{2-3}
                                    & $2 \times 2$ & $0.832 \pm 0.02$ \\
        \cline{2-3}
                                    & $7^1$        & $\mathbf{0.833 \pm 0.02}$ \\
        \hline
        \multirow{3}{*}{MNIST}      & $3 \times 3$ & $0.993 \pm 0.01$ \\
        \cline{2-3}
                                    & $2 \times 2$ & $\mathbf{0.997 \pm 0.01}$ \\
        \cline{2-3}
                                    & $7^1$        & $0.995 \pm 0.01$ \\
        \hline
    \end{tabular}

    \caption{S-CNN and H-CNN test results after 100 epochs of training as averaged over five runs.}
    \label{table:CNN_test_results}
\end{table}

\subsection{Square and hexagonal convolutional neural networks}

%%%%%%%%%%%%%%%%%%%%%%%%%%%%%%%%%%%%% Text %%%%%%%%%%%%%%%%%%%%%%%%%%%%%%%%%%%%%
To enable the comparison of S-CNN and H-CNN models, we deployed different pooling sizes and tested them against various data sets after transformation with an initial image resolution of $32 \times 32$ and $34 \times 30$ pixels respectively. These include the pooling sizes of $2 \times 2$, $3 \times 3$, and hexagonal kernels of size $7^1$. The therefore realized S-CNN and H-CNN architecture configurations are inspired by \textit{LeCun et al.'s} LeNet-5 \cite{LeCun1998} with additionally three Dropout layers \cite{Hinton2012} as shown in Tables \ref{table:S-CNN_model_configuration} and \ref{table:H-CNN_model_configuration} for 100 classes, resulting in $565\thinspace 956$ and $363\thinspace 012$ trainable parameters respectively. However, deploying a square lattice based pooling kernel of size $3 \times 3$ following the given S-CNN layer configuration results in $336\thinspace 580$ trainable parameters with a mere $(6, 6, 64)$ output shape for layer \textit{pool} with $64$ filters.

For training itself, the Glorot initializer \cite{Glorot2010} for weight initialization as well as the Adam optimizer \cite{Kingma2015} with standard learning rate and a batch size of $32$ are hereinafter deployed for all S-CNN- and H-CNN-based models.

These were then evaluated against the CINIC-10 \cite{Darlow2018}, COIL-100 \cite{Nene1996b}, and MNIST \cite{LeCun1998} data sets. Table~\ref{table:CNN_test_results} shows the test results for all three CNN configurations as averaged over five runs with a data set split ratio as given by the respective data sets. While the hexagonal pooling layer itself already results in a slight shift and rotation given by the offset mapping based on the different sub-Hexarray orders, we conducted our tests without any additional data augmentation and included a distorted version of the COIL-100 data set, hereinafter also referred to as D-COIL(-100). D-COIL was created by augmenting every image, including randomized rotations, translations, scaling, reflection, and elastic and perspective transformations\footnote{\url{https://github.com/aleju/imgaug}}.

Whereas a comparison of all test results for the original versions of the data sets reveals increased accuracies for the S-CNN with a pooling size of $3 \times 3$, the results for D-COIL show an increased test accuracy for the H-CNN. Furthermore, the reduction of trainable parameters also enables the reduction of the CNN's complexity by switching to its hexagonal counterpart, whereas hexagonal layers can improve train and test rates while only resulting in relatively minor parameter trade-offs. To consider their training behavior, we included the training results as shown in Fig.~\ref{figure:CNN_test_results}. These display a slower learning rate for the H-CNN on the original data sets, whereas the results for D-COIL suggest a higher adaptability when training with highly augmented data.
%%%%%%%%%%%%%%%%%%%%%%%%%%%%%%%%%%%%% Text %%%%%%%%%%%%%%%%%%%%%%%%%%%%%%%%%%%%%

\subsection{Transformation efficiency}

%%%%%%%%%%%%%%%%%%%%%%%%%%%%%%%%%%%%% Text %%%%%%%%%%%%%%%%%%%%%%%%%%%%%%%%%%%%%
To quantify the transformation quality of the square and the hexagonal lattice format based transformation, the following so-called transformation efficiencies can be obtained by comparing the resulting transformed image to the original one. Given the original image $I$ and its transformed image $K$, the absolute error can be determined by weighting the from projection of $K$ onto $I$ obtained subareas $a \in A$ in sub-pixel resolution, whereas the subareas area is given by $|a|$. The computation of the mean squared error (MSE) is shown in \eqref{equation:MSE}.
%%%%%%%%%%%%%%%%%%%%%%%%%%%%%%%%%%%%% Text %%%%%%%%%%%%%%%%%%%%%%%%%%%%%%%%%%%%%

\begin{equation}
    \mathit{MSE} = \frac{1}{|A|} \cdot \sum\limits_{a \in A} |a| \cdot [I(a) - K(a)]^2
    \label{equation:MSE}
\end{equation}

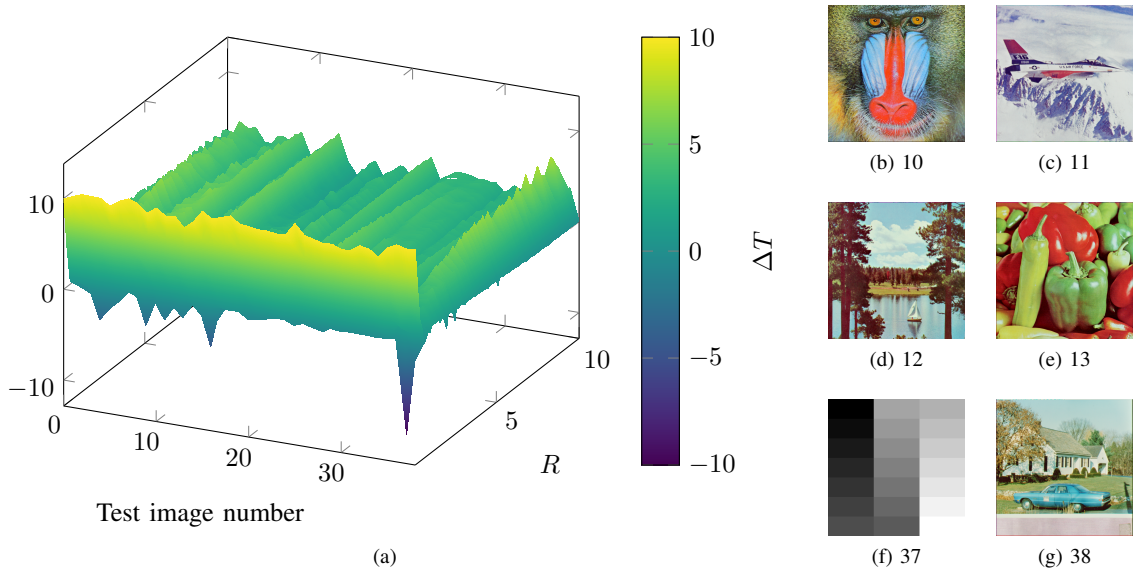
\begin{figure*}[tb]
    \centering

    \begin{tabular}{cc}
        \subfloat[]{
            \begin{tikzpicture}
                \begin{axis}[
                 colormap/viridis, colorbar, mesh/cols=39, point meta min=-10, point meta max=10,
                 xlabel={Test image number}, ylabel={$R$}, colorbar style={ylabel={$\Delta T$}}]
                    \addplot3[surf, shader=interp] file {data/transformation_efficiency.csv};
                \end{axis}
            \end{tikzpicture}} &
            \begin{tabular}[b]{cc}
                \subfloat[10]{\includegraphics[width=0.1\textwidth]{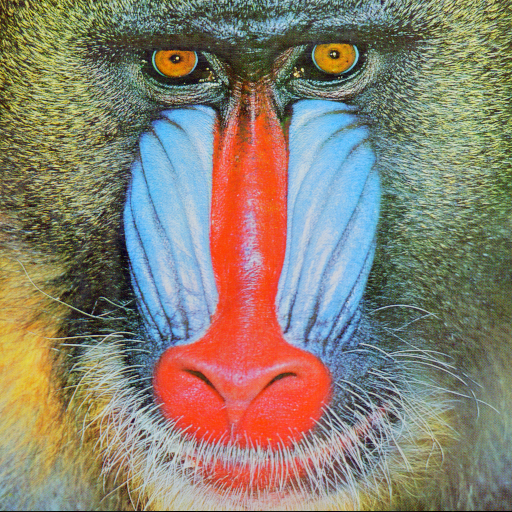}} &
                    \subfloat[11]{\includegraphics[width=0.1\textwidth]{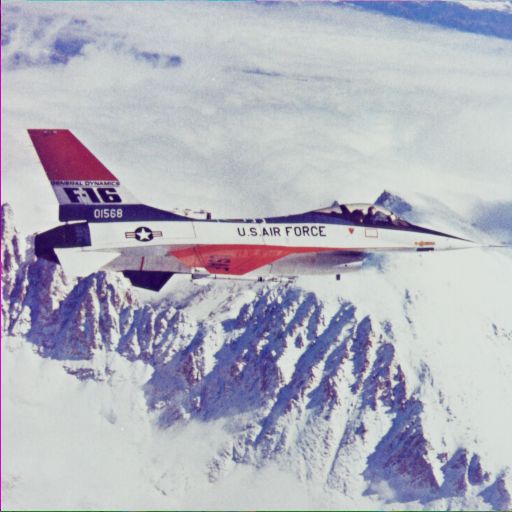}} \\
                \subfloat[12]{\includegraphics[width=0.1\textwidth]{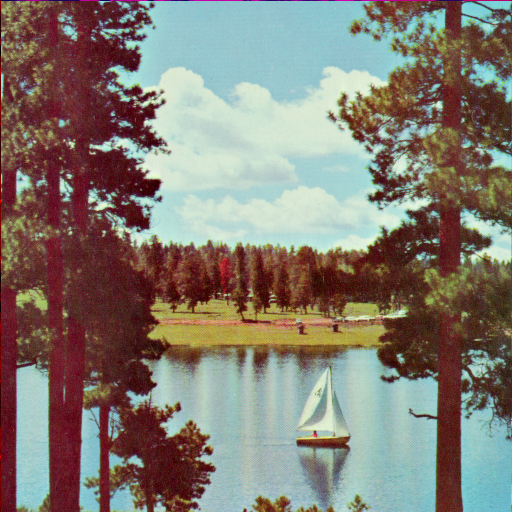}} &
                    \subfloat[13]{\includegraphics[width=0.1\textwidth]{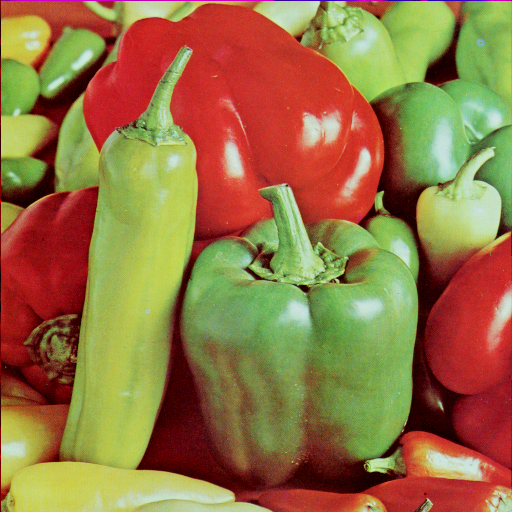}} \\
                \subfloat[37]{\includegraphics[width=0.1\textwidth]{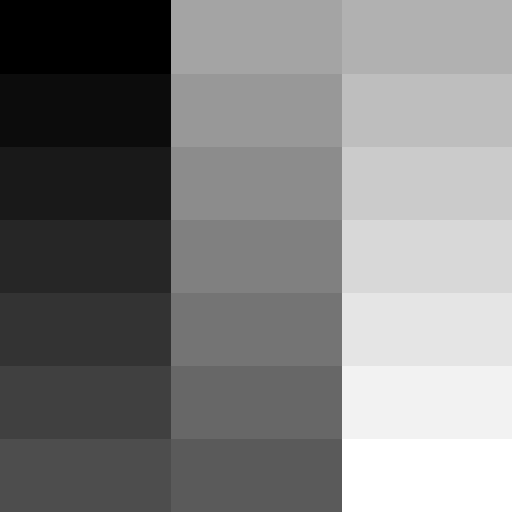}} &
                    \subfloat[38]{\includegraphics[width=0.1\textwidth]{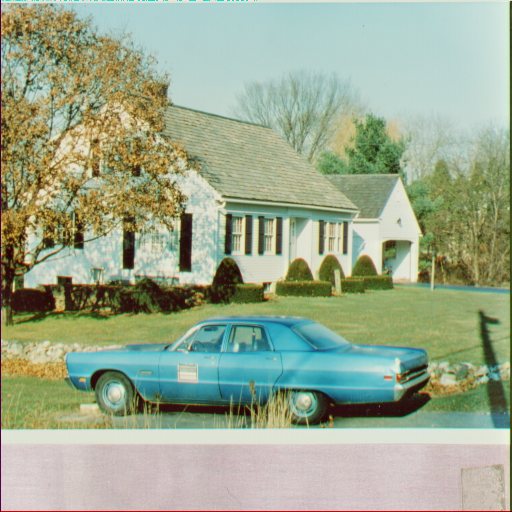}} \\
            \end{tabular}
    \end{tabular}

    \caption{Square and hexagonal image transformation in comparison using the USC-SIPI image database \cite{Weber1993}.}
    \label{figure:T_test_results}
\end{figure*}

%%%%%%%%%%%%%%%%%%%%%%%%%%%%%%%%%%%%% Text %%%%%%%%%%%%%%%%%%%%%%%%%%%%%%%%%%%%%
For the differences of square ($T_q$) and hexagonal image transformation ($T_h$), hereinafter also referred to as transformation efficiency $T$, the transformation efficiency differences ($\Delta T = T_h - T_q$) are shown in Fig.~\ref{figure:T_test_results} using the peak signal-to-noise ratio (PSNR) in \eqref{equation:PSNR}. These results were obtained with the USC-SIPI image database \cite{Weber1993}. Each Hexint was then interpolated with its circumradius $R$, whereas the square lattice based transformation is based on the resulting Hexarray resolution as the square image target resolution.
%%%%%%%%%%%%%%%%%%%%%%%%%%%%%%%%%%%%% Text %%%%%%%%%%%%%%%%%%%%%%%%%%%%%%%%%%%%%

\begin{equation}
    \mathit{PSNR} = 10 \cdot \log_{10}\left(\frac{\mathit{MAX}_I^2}{\mathit{MSE}}\right)
    \label{equation:PSNR}
\end{equation}

%%%%%%%%%%%%%%%%%%%%%%%%%%%%%%%%%%%%% Text %%%%%%%%%%%%%%%%%%%%%%%%%%%%%%%%%%%%%
Following \textit{Field} \cite{Field1987}, the test images which are most commonly characterized as natural images (b--e) show evidently increased transformation efficiencies. A counter example can be observed for the test images number 37 (f) and 38 (g), which can't a priori be reliably approximated based on the hexagonal lattice format due to the large number of occurring vertical and horizontal structures. Furthermore, near the circumradius of $1$, a decline in $T_h$ can be observed. This can be explained in particular by the resulting square pixel side length, which reaches the image resolution of the original image, therefore resulting in only minor interpolation artifacts.
%%%%%%%%%%%%%%%%%%%%%%%%%%%%%%%%%%%%% Text %%%%%%%%%%%%%%%%%%%%%%%%%%%%%%%%%%%%%

\subsection{Performance}

%%%%%%%%%%%%%%%%%%%%%%%%%%%%%%%%%%%%% Text %%%%%%%%%%%%%%%%%%%%%%%%%%%%%%%%%%%%%
As previous research often omits the evaluation for performance of the hexagonal transformation itself, we conducted an evaluation of the most commonly deployed SAA-based addressing scheme. Due to the fact that square lattice based architectures often tend to be based on either one-dimensional linewise or two-dimensional addressing schemes, they also greatly reduce the locality for hexagonal transformations, as their locality differs, and are therefore outperformed by their square counterparts.

The hexagonal image processing framework HIP, introduced by \textit{Middleton and Sivaswamy} \cite{Middleton2005}, is one of the few publicly available frameworks for hexagonal image processing. HIP performs its hexagonal image transformation through an iterative traversal of the septree resulting from Fig.~\ref{figure:addressing_schemes}b, where neighborhood relationships across the sub-Hexarrays are computed at each step of the transformation. Table~\ref{table:T_runtime_test_results} shows a comparison with Hexnet for the Hexarrays with a total pixel count of equal to the orders 5 to 7. These test results where obtained using the USC-SIPI image database with an original image resolution of $128 \times 128$ pixels with bilinear interpolation and averaged over five runs.

The resulting addressing scheme benefits are indicated by the reduced runtimes, ranging from factor $2\thinspace 500$ to $4\thinspace 000$ over HIP. Furthermore, the hexagonal transformation reaches close to square baseline runtimes and can therefore be seen as a suitable initial processing step.\footnote{CPU: Intel Core i7-6500U~@ 2.50GHz, BogoMips: 5184, CPU load: ca. 99~\% (one core), VmPeak: ca. 814~MiB (max. for order 7)}
%%%%%%%%%%%%%%%%%%%%%%%%%%%%%%%%%%%%% Text %%%%%%%%%%%%%%%%%%%%%%%%%%%%%%%%%%%%%

\begin{table}[tb]
    \renewcommand{\arraystretch}{1.2}
    \centering

    \begin{tabular}{|M{2cm}|M{1.5cm}|M{1.5cm}|M{1.5cm}|}
        \hline
        Pixel \# & $7^5$ & $7^6$ & $7^7$ \\
        \hline
        \noalign{\vskip 2pt}

        \hline
        HIP             & $0.167 \pm 0.04$         & $0.033 \pm 0.01$         & $0.006 \pm 0.0$ \\
        \hline
        \textbf{Hexnet} & $\mathbf{666.7 \pm 2.1}$ & $\mathbf{111.1 \pm 0.9}$ & $\mathbf{14.93 \pm 0.1}$ \\
        \hline
        \noalign{\vskip 2pt}

        \hline
        Square baseline & $675 \pm 0.19$ & $125 \pm 0.4$ & $17.5 \pm 0.1$ \\
        \hline
    \end{tabular}

    \caption{Hexagonal image processing frameworks transformations and square lattice format baseline transformation performance comparison in images per second as averaged over five runs.}
    \label{table:T_runtime_test_results}
\end{table}

\section{Summary and outlook}

%%%%%%%%%%%%%%%%%%%%%%%%%%%%%%%%%%%%% Text %%%%%%%%%%%%%%%%%%%%%%%%%%%%%%%%%%%%%
The in this contribution proposed hexagonal deep learning framework provides the foundation of a general application-oriented approach for hexagonal image processing in the context of machine learning. The introduced framework, called Hexnet, serves to synthesize the advantages of the domains of biologically inspired hexagonal image processing with machine learning approaches by utilizing hexagonal artificial neural networks.

Hexnet implements its hexagonal image transformation as based on its hexagonal architecture, which is combining the advantages of classical square as well as hexagonal lattice format based addressing schemes. Following this addressing scheme, the introduced hexagonal deep neural networks, also called H-DNN, include not only hexagonal versions for convolutional and pooling layers, but also the foundation for more sophisticated hexagonal models.

The results of our created test environment show that H-DNNs can outperform conventional DNNs, while an increase of training and test rates show the possible benefits of their hexagonal pendants. The hexagonal image transformation itself as an initial processing step achieves a convincing transformation result, and even its performance can compare to its square baseline. The implemented functionality as of the current state of art of recording and output devices demonstrates the feasibility and applicability of hexagonal deep learning, whereas the realized processing steps as well as layers and models can be efficiently implemented in already existing ones by deploying the provided functionality.

Subsequent machine learning models can be based on this framework and form the hexagonal equivalent to conventional deep neural networks. However, Hexnet still has to be deployed under real world test conditions, as more novel and application-specific models have to developed and investigated.
%%%%%%%%%%%%%%%%%%%%%%%%%%%%%%%%%%%%% Text %%%%%%%%%%%%%%%%%%%%%%%%%%%%%%%%%%%%%

\section*{Acknowledgment}

%%%%%%%%%%%%%%%%%%%%%%%%%%%%%%%%%%%%% Text %%%%%%%%%%%%%%%%%%%%%%%%%%%%%%%%%%%%%
The European Union and the European Social Fund for Germany partially funded this research.
%%%%%%%%%%%%%%%%%%%%%%%%%%%%%%%%%%%%% Text %%%%%%%%%%%%%%%%%%%%%%%%%%%%%%%%%%%%%

\bibliographystyle{IEEEtran_Tobias}
\bibliography{library}

\end{document}